\documentclass[runningheads]{llncs}
\usepackage[T1]{fontenc}
\usepackage{graphicx,verbatim}
\usepackage[hidelinks]{hyperref}
\usepackage{color}

\usepackage{float}
\usepackage{booktabs}
\usepackage{xurl}
\usepackage{longtable}
\usepackage{amsmath,amssymb}
\usepackage{bbding}

\usepackage{microtype}
\begin{document}

\title{EyeVQA: Benchmarking Ophthalmic Vision-Language Models from Recognition to Spatial Grounding}
\titlerunning{EyeVQA}

\author{
Gujie Shao\textsuperscript{$\star$}, 
Zixun Xie\textsuperscript{$\star$}, 
Xuechun Xing\textsuperscript{$\star$}, 
Ruixiang Wang\textsuperscript{$\star$}, 
Ziyun Lan, \\
Yanlin Qi,
Gangyi Zhang, 
Yuxin Yang, 
Dawei Li\Envelope, 
Haiming Tang\Envelope
}

\authorrunning{G. Shao et al.}

\institute{
\email{haiming@comp.nus.edu.sg}, \email{lidawei@pku.edu.cn}
\\
\textsuperscript{$\star$}Equal contribution \quad \Envelope Corresponding authors
}
  
\maketitle
\begin{abstract}
Vision-language models (VLMs) have shown increasing potential for medical image understanding, yet their capabilities in ophthalmic imaging remain insufficiently characterized. 
Existing ophthalmic datasets are typically designed for individual diseases or specialized tasks, making it difficult to systematically evaluate whether VLMs can move beyond disease recognition toward comparative reasoning and fine-grained spatial grounding. We introduce EyeVQA, a unified visual question answering benchmark for comprehensive evaluation of ophthalmic VLMs. 
EyeVQA is constructed from 21 available ophthalmic datasets and contains 20,000 question–answer pairs spanning six disease groups and seven question types: Single-Choice, Multi-Select, Variable-Select, True-False, Ranking, Point Location, and Bounding Box. 
Gold answers are deterministically derived from source-provided diagnoses, severity grades, clinical findings, segmentation masks, bounding boxes, and anatomical landmarks, enabling reproducible evaluation without relying on model-generated annotations. 
Notably, 44.5\% of the questions require reasoning across multiple images, extending evaluation beyond conventional single-image medical VQA. We benchmark fourteen representative general-purpose, scientific, and medically specialized VLMs under a unified zero-shot protocol. 
The best-performing model only achieves an overall score of 62.8, while substantial gaps remain in spatial grounding and cross-task generalization. 
These results highlight the limitations of current VLMs in comprehensive ophthalmic visual understanding and establish EyeVQA as a diagnostic benchmark for developing more reliable and spatially grounded ophthalmic multimodal models. The project page is available at \url{https://github.com/PKUTHM/EyeVQA}.

\end{abstract}

\section{Introduction}
Recent breakthroughs in vision-language models (VLMs)~\cite{bai2025qwen3,xu2025lingshu,hong2025glm,internlm2026interns2preview35b,sellergren2026medgemma} have marked a paradigm shift in multimodal artificial intelligence by integrating visual perception with complex natural-language reasoning. In medical imaging, this interactive paradigm is particularly transformative: clinical diagnosis inherently demands more than passive pattern recognition; it requires linking subtle pathological signs with medical concepts, performing cross-image comparative reasoning, and verifying fine-grained spatial evidence to support diagnostic conclusions.

Ophthalmic imaging, particularly fundus photography, serves as a paramount domain for evaluating multi-level visual understanding in multimodal AI. However, existing ophthalmic image analysis resources remain fragmented~\cite{zhou2023foundation,xie2026fine,silva2025foundation}. Most available datasets were built for isolated tasks—such as diabetic retinopathy grading~\cite{eyepacs}, glaucoma screening~\cite{li2019attention}, or pathological myopia diagnosis~\cite{pachade2021retinal}, offering insufficient disease diversity or task flexibility to holistically assess a VLM's clinical reasoning capabilities. Moreover, evaluation restricted to single-image classification fails to reflect real-world clinical workflows, where ophthalmologists routinely compare longitudinal images or bilateral fundus scans.

Visual question answering (VQA) provides a versatile framework for multimodal evaluation~\cite{wang2026towards}. Nevertheless, current medical VQA paradigms suffer from three major limitations when applied to ophthalmology:
\begin{enumerate}
    \item \textbf{Overemphasis on Single-Image Semantics:} Existing benchmarks primarily focus on single-image semantic identification or global disease classification, overlooking complex clinical reasoning patterns.
    \item \textbf{Lack of Cross-Image Relational Reasoning:} Standard single-image evaluation cannot assess a model's ability to perform comparative analysis, such as ordinal severity assessment or anatomical progression across multiple visual inputs.
    \item \textbf{Absence of Explicit Spatial Grounding Verification:} Semantic correctness alone does not guarantee that a VLM grounds its conclusions on correct anatomical evidence, making fine-grained spatial localization crucial for trustworthy medical AI.
\end{enumerate}

\begin{figure}[t]
    \centering
    \includegraphics[width=0.8\textwidth]{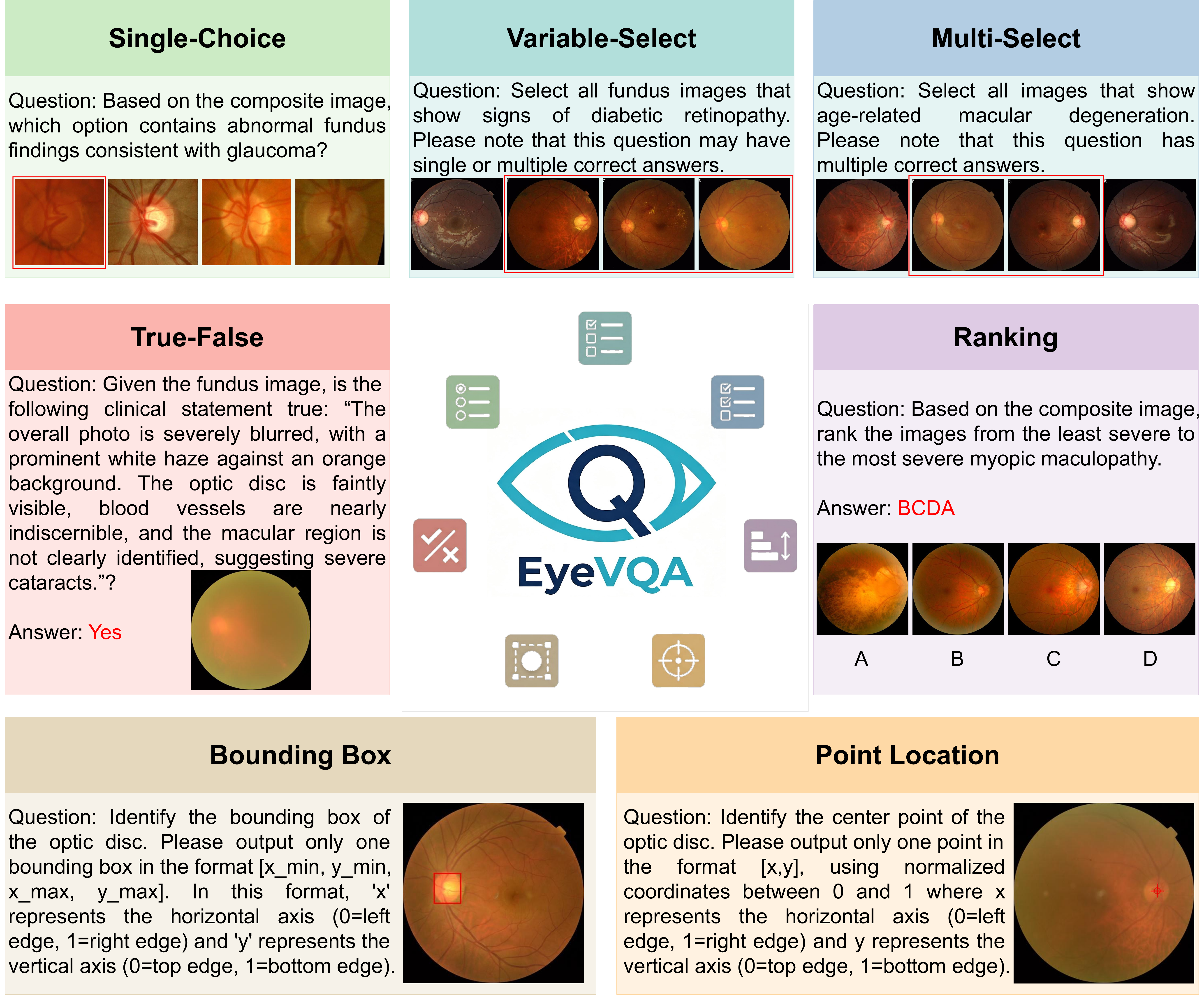}
    \caption{Overview of the EyeVQA benchmark.}
    \label{fig:overview}
\end{figure}

To bridge these gaps, we present \textbf{EyeVQA}, a unified benchmark specifically designed to evaluate the multi-faceted reasoning and spatial grounding capabilities of VLMs in ophthalmology, as shown in Fig.~\ref{fig:overview}. EyeVQA aggregates 21 ophthalmic datasets into 20,000 clinically grounded question--answer (QA) pairs across six primary disease categories: Diabetic Retinopathy (DR), Glaucoma (GL), Pathological Myopia (PM), Macula Disease (MD), Multiple (MU), and Other (OT). 

To systematically evaluate complementary reasoning dimensions, EyeVQA formulates seven standardized question types: Single-Choice (SC), Multi-Select (MS), Variable-Select (VS), True-False (TF), Ranking (RK), Point Location (PL), and Bounding Box (BB). Together, these question formats span core clinical problems, including disease presence identification, severity grading, image--description correspondence, comparative disease progression, and precise anatomical or lesion grounding. Crucially, all gold answers and spatial targets are deterministically derived from expert annotations rather than generated by secondary LLMs, eliminating hallucinated references. Furthermore, EyeVQA incorporates 10,344 four-panel composites, making \textbf{44.5\%} of the benchmark (8,902 questions) require comparative reasoning across multiple images---a key setting absent from prior medical VQA benchmarks.

We conduct a standardized evaluation of fourteen representative VLMs spanning open-source general-purpose, close-source general-purpose, scientific, and medically specialized model families under a zero-shot, inference-only protocol. Our experimental findings reveal crucial insights into current multimodal architectures:
\begin{itemize}
    \item Current state-of-the-art VLMs still fall short of robust clinical utility: the leading model, Intern-S2-Preview-397B, achieves an overall performance of only 62.8\%.
    \item High semantic accuracy does not guarantee spatial grounding precision, as models often excel at verbal disease recognition but fail significantly at coordinate-based localization tasks (PL and BB).
    \item Specialized medical VLMs do not consistently outperform top-tier generalist or scientific models, underscoring the need for balanced training on both domain knowledge and precise spatial alignment.
\end{itemize}

Our main contributions are summarized as follows:
\begin{enumerate}
    \item \textbf{A Large-Scale, Unified Benchmark:} We introduce EyeVQA, unifying 21 datasets into 20,000 QA pairs across six disease groups and seven standardized question types, establishing a multi-dimensional testbed from visual perception to clinical reasoning.
    \item \textbf{Deterministic Grounding \& Multi-Image Design:} All QA pairs are deterministically constructed from verified expert annotations, with 44.5\% of the benchmark featuring four-panel composite images to evaluate cross-image comparative reasoning.
    \item \textbf{Extensive Benchmarking \& Diagnostic Insights:} We systematically evaluate fourteen leading general, scientific, and medical VLMs under unified zero-shot constraints, uncovering critical bottlenecks in fine-grained visual grounding and cross-modal alignment.
\end{enumerate}
\section{Construction of EyeVQA}

\subsection{Dataset Collection}
To comprehensively evaluate ophthalmic vision-language models (VLMs), we collected \textbf{21 available ophthalmic datasets} covering major retinal conditions across diverse platforms, such as \textit{Scientific Data}, Kaggle, Hugging Face, Zenodo, and GitHub. These datasets exhibit rich visual and textual heterogeneity, ranging from disease classification and lesion segmentation to clinical descriptions and paired multimodal imaging. A comprehensive description of the data sources, annotation formats, and standardization procedures is detailed in Section~\ref*{supp:dataset_collection} in the Supplementary Material (with a complete overview summarized in Table~\ref*{tab:datasets}).

\subsection{Question--Answer Design}
\label{subsec:qa_design}

Following established benchmark designs in general and medical VQA~\cite{liu2025human,lau2018dataset,he2020pathvqa}, EyeVQA formulates seven standardized question types spanning diverse reasoning granularities, as illustrated in Fig.~\ref{fig:overview}. 

All QA pairs are deterministically constructed from verifiable ground truths, including diagnostic labels, severity grades, clinical findings, segmentations, and landmark annotations. 
Specifically, \textbf{SC}, \textbf{MS}, and \textbf{VS} evaluate diagnostic recognition and candidate selection with varying answer cardinalities. 
\textbf{TF} verifies clinical statement correctness, while \textbf{RK} assesses ordinal disease progression or anatomical metrics across composite images. 
For spatial grounding, \textbf{PL} and \textbf{BB} require precise coordinate predictions for clinical landmarks and lesion areas. 

Detailed prompt templates, distractor sampling strategies, and coordinate transformation specifications are deferred to Section~\ref*{supp:qa_details} in the Supplementary Material.

\subsection{Statistics and Analysis}
\label{subsec:statistics}

\subsubsection{Scale and Composition.}
EyeVQA contains \textbf{20{,}000} standardized question--answer pairs referencing \textbf{27{,}935} fundus images from 21 source datasets. Beyond single-image VQA, it incorporates 10{,}344 four-panel composite images, making \textbf{8{,}902} questions (44.5\%) require multi-image relational reasoning. 
The benchmark spans seven standardized question types (SC, MS, VS, TF, RK, PL, BB) and six disease groups (DR, GL, PM, MD, OT, MU), ensuring comprehensive coverage from visual perception to multi-disease reasoning (Fig.~\ref{fig:dataset_overview}).

\subsubsection{Text Characteristics and Balanced Options.}
\begin{figure}[!htbp]
  \centering
  \includegraphics[width=\linewidth]{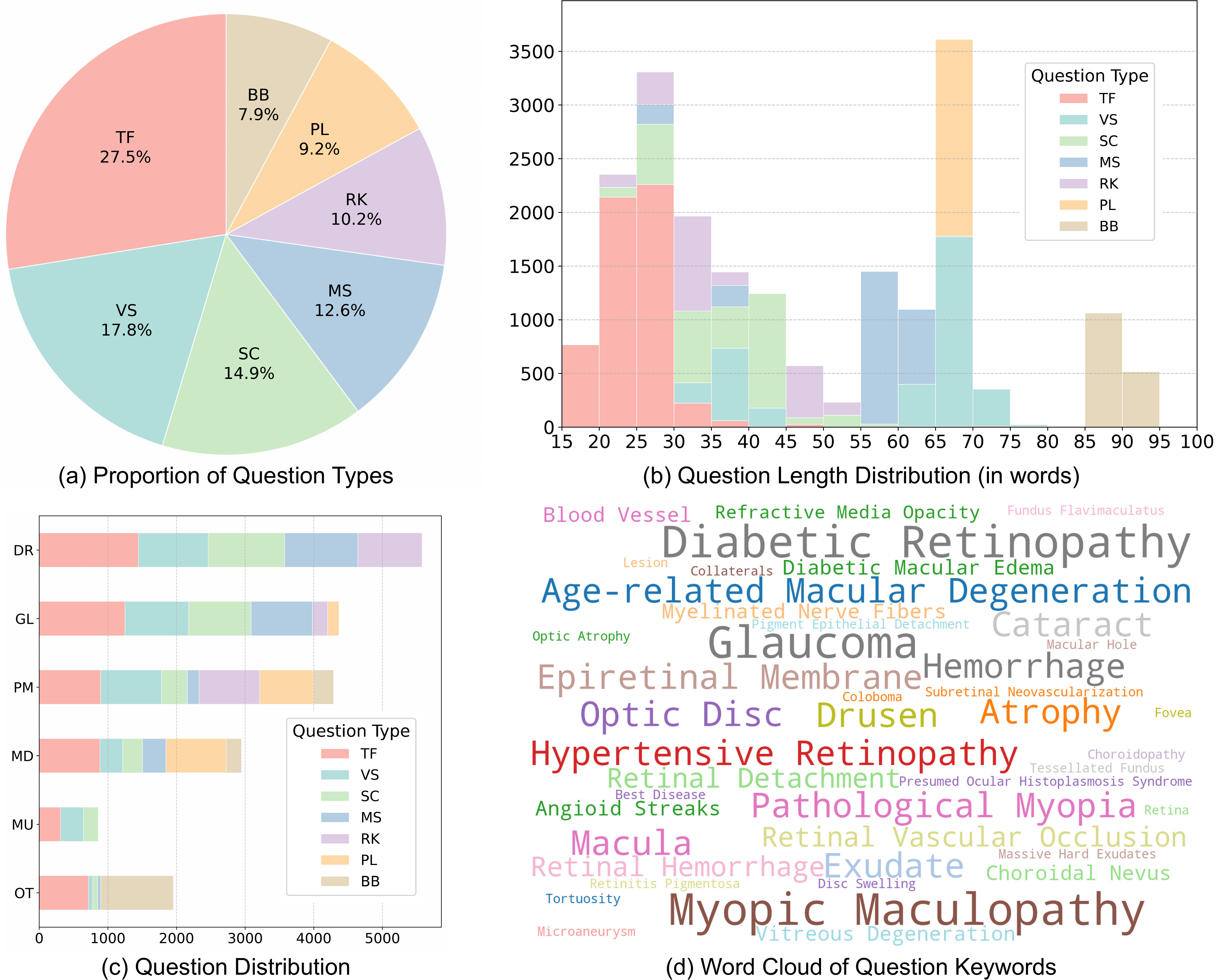}
  \caption{Dataset overview of EyeVQA: 
  (a) distribution of question types; 
  (b) question length distribution by type; 
  (c) question-type composition within each disease group; 
  (d) clinical vocabulary frequency over the 20{,}000 questions.}
  \label{fig:dataset_overview}
\end{figure}

Question lengths average \textbf{46.0} words (median 40, range 16--170), exhibiting a multi-modal distribution aligned with task output complexity (Fig.~\ref{fig:dataset_overview}b): TF prompts are the most concise (mean 24.3 words), whereas spatial grounding tasks (e.g., BB, mean 89.7 words) require explicit coordinate specifications. 
To prevent shortcut learning and language priors, options are carefully balanced across tasks: SC options are nearly uniform (A: 24.2\%, B: 24.7\%, C: 26.6\%, D: 24.5\%), TF maintains a well-balanced binary distribution, and MS/VS cover diverse answer cardinalities.

\section{Experiments}

\subsection{Experimental Setting}

We evaluate fourteen representative vision-language models (VLMs), spanning open-source general-purpose, closed-source general-purpose, science-oriented, and medically specialized model families. The open-source general-purpose group comprises GLM-4V-Flash \cite{zhipuai2026glm4vflash}, GLM-4.1V-Thinking-Flash \cite{hong2025glm,zhipuai2026glm41vflash}, GLM-4.6V-Flash \cite{zhipuai2026glm46vflash}, Qwen2.5-VL-3B-Instruct and Qwen2.5-VL-7B-Instruct \cite{Qwen2.5-VL}, and Qwen3-VL-8B-Instruct \cite{bai2025qwen3}, while the closed-source general-purpose group includes Gemini-3.1-Flash-Lite \cite{gemini3.1} and Gemini-3.5-Flash-Lite \cite{gemini3.5}. The science-oriented group contains Intern-S1-Pro, a trillion-parameter mixture-of-experts model \cite{zou2026intern}, together with Intern-S2-Preview-35B \cite{internlm2026interns2preview35b} and Intern-S2-Preview-397B \cite{internlm2026interns2preview397b}. The medical group includes Lingshu-7B \cite{xu2025lingshu}, MedGemma-4B-IT \cite{sellergren2025medgemma}, and its successor MedGemma-1.5-4B-IT \cite{sellergren2026medgemma}. This selection enables comparisons across model scales and between general-purpose, science-oriented, and medically specialized VLMs.

All models are evaluated in a zero-shot, inference-only setting without training or fine-tuning. Open-source VLMs are deployed with vLLM \cite{kwon2023efficient}, while API-based models are accessed through their official services. Each model receives the same question and standardized image input. Task-specific format-enforcing prompts require concise structured responses, which are parsed with regular expressions to extract the final answers. This action minimizes the bias introduced by the generation style of VLMs.

\subsection{Evaluation Metrics}
We evaluate model performance across seven distinct question types: Single-Choice ($\mathrm{SC}$), True-False ($\mathrm{TF}$), Multi-Select ($\mathrm{MS}$), Variable-Select ($\mathrm{VS}$), Ranking ($\mathrm{RK}$), Bounding Box ($\mathrm{BB}$), and Point Location ($\mathrm{PL}$). Detailed mathematical formulations for each metric—including exact-match accuracy, Hamming metric, Kendall's Tau, Intersection over Union (IoU), and Euclidean distance tolerance—are provided in Section~\ref*{supp:evaluation_metrics} in the Supplementary Material. For cross-task comparison, all raw metric scores are linearly normalized to $[0, 100]$, and refused responses are assigned a score of zero.

\subsection{Results}

\subsubsection{Disease-category results}
Table~\ref{tab:main_results} compares performance across the six disease categories. Intern-S2-397B achieves the highest overall score of 62.8 and leads in five categories, including DR, GL, PM, MD, and MU, supported by its particularly strong performance on MD (73.4). Intern-S2-35B follows with the second-highest overall score of 59.3 and remains competitive across disease categories, while Gemini-3.1-Flash-Lite secures the top rank in the OT category with 73.4 (surpassing Intern-S2-35B which ranks second with 68.1). Among the medically specialized models, MedGemma-4B performs strongly on DR, ranking second with 62.5, whereas Lingshu-7B and MedGemma-1.5-4B obtain more moderate overall scores. These results indicate that the Intern-S2 models provide the most consistent cross-disease performance, while lightweight closed-source models deliver exceptional category-specific results.

\subsubsection{Question-type results}
The question-type breakdown reveals complementary model strengths. Intern-S2-397B ranks first on SC, VS, TF, and PL, and is only 0.3 points behind the best result on MS. Intern-S2-35B leads on MS, while nearly matching Intern-S2-397B on SC and VS. In contrast, Intern-S1-Pro achieves the highest RK score (60.0) despite its relatively low BB score (17.0), suggesting strong ordering ability but limited fine-grained spatial grounding. Qwen3-VL-8B is another notable exception: although its overall score is 48.9, it ranks third on BB with 46.2. MedGemma-4B attains the second-best TF score (71.1) but performs poorly on BB (6.2), further showing that recognition and spatial localization capabilities do not necessarily improve together. Furthermore, the lightweight closed-source models exhibit remarkable spatial and grounding capabilities, with Gemini-3.1-Flash-Lite achieving the highest Bounding Box (BB) score of 58.1 and Gemini-3.5-Flash-Lite securing the second-best Point Location (PL) score of 65.3, highlighting their strong potential in fine-grained visual localization tasks.

\begin{table}[htbp]
\centering
\caption{Performance comparison of the evaluated VLMs. \textbf{Bold} indicates the best result, and \underline{Underline} indicates the second-best result.}
\label{tab:main_results}
\begingroup
\setlength{\tabcolsep}{1.25pt}
\renewcommand{\arraystretch}{1.05}
\fontsize{8}{9}\selectfont
\begin{tabular}{@{}l|c|*{6}{c}|*{7}{c}@{}}
\toprule
Model & Total & DR & GL & PM & MD & MU & OT & SC & MS & VS & TF & RK & PL & BB \\
\midrule
GLM-4V-Flash & 39.3 & 37.5 & 45.0 & 41.6 & 41.5 & 35.7 & 25.4 & 37.1 & 48.9 & 31.8 & 42.1 & 53.2 & 51.3 & 4.0 \\
GLM-4.1V-Flash & 49.5 & 49.1 & 46.3 & 46.9 & 55.5 & 46.6 & 55.7 & 43.1 & 51.2 & 38.5 & 60.3 & 57.3 & 50.6 & 34.5 \\
GLM-4.6V-Flash & 45.6 & 44.8 & 44.0 & 44.6 & 45.2 & 38.5 & 57.3 & 37.6 & 54.0 & 36.4 & 51.6 & 57.4 & 27.9 & \underline{51.8} \\
Intern-S1-Pro & 54.1 & 58.7 & 51.9 & 51.9 & 58.6 & 54.0 & 44.4 & 49.6 & 59.0 & 44.5 & 70.2 & \textbf{60.0} & 50.7 & 17.0 \\
Intern-S2-35B & \underline{59.3} & 60.0 & \underline{57.1} & \underline{55.6} & 60.8 & \underline{60.4} & \underline{68.1} & \underline{60.1} & \textbf{63.3} & \underline{50.5} & 70.5 & \underline{59.1} & 45.2 & 49.4 \\
Intern-S2-397B & \textbf{62.8} & \textbf{63.2} & \textbf{57.9} & \textbf{57.9} & \textbf{73.4} & \textbf{62.5} & 67.7 & \textbf{60.2} & \underline{63.0} & \textbf{51.6} & \textbf{76.2} & 55.8 & \textbf{70.7} & 45.5 \\
Lingshu-7B & 50.9 & 58.6 & 46.3 & 48.6 & 52.2 & 47.7 & 43.8 & 53.0 & 59.5 & 45.0 & 61.8 & 56.8 & 35.1 & 19.2 \\
MedGemma-1.5-4B & 49.9 & 59.8 & 53.3 & 40.3 & 51.0 & 56.0 & 31.1 & 48.7 & 57.8 & 40.8 & 64.6 & 50.0 & 51.3 & 7.5 \\
MedGemma-4B & 51.5 & \underline{62.5} & 52.2 & 47.2 & 48.7 & 57.0 & 30.1 & 47.6 & 53.6 & 41.9 & \underline{71.1} & 58.1 & 46.5 & 6.2 \\
Qwen2.5-VL-3B & 41.0 & 44.4 & 34.4 & 43.5 & 38.2 & 54.3 & 39.0 & 39.2 & 51.2 & 39.7 & 45.9 & 52.1 & 23.2 & 20.1 \\
Qwen2.5-VL-7B & 47.1 & 45.5 & 49.4 & 48.4 & 51.1 & 54.9 & 34.6 & 45.2 & 58.2 & 44.2 & 54.4 & 54.8 & 44.6 & 7.3 \\
Qwen3-VL-8B & 48.9 & 46.9 & 50.8 & 48.5 & 40.7 & 47.5 & 64.7 & 44.7 & 58.8 & 39.4 & 60.1 & 49.5 & 29.0 & 46.2 \\
Gemini-3.1-Flash & 57.0 & 55.7 & 49.5 & 53.0 & 66.8 & 54.1 & \textbf{73.4} & 53.4 & 58.6 & 44.3 & 67.3 & 54.6 & 56.5 & \textbf{58.1} \\
Gemini-3.5-Flash & 55.9 & 55.2 & 55.4 & 48.4 & \underline{69.5} & 53.3 & 56.5 & 53.1 & 58.9 & 45.1 & 70.2 & 49.8 & \underline{65.3} & 27.8 \\
\bottomrule
\end{tabular}
\endgroup
\end{table}

\subsection{Key Findings}

We summarize four key findings from the benchmark, as supported by Table~\ref{tab:main_results}.

\subsubsection{Larger models perform better in most tasks.}
Performance generally increases with model scale across the benchmark. The scaling trend is clearest within matched model families and is also reflected in the overall ranking, where larger models achieve stronger aggregate results and broader performance across disease categories and question types.

\subsubsection{Medical specialization offers parameter-efficient benefits.}
Compact medically specialized VLMs achieve competitive performance relative to substantially larger science-oriented models across clinically relevant recognition, reasoning, and localization tasks. This recurring trend indicates that domain-specific medical knowledge improves the effective use of limited model capacity.

\subsubsection{Reasoning and spatial grounding are partially decoupled.}
Across the evaluated models, performance rankings on recognition and ordering tasks show limited correspondence with rankings on bounding-box and point-localization tasks. The systematic difference between these task groups indicates that semantic reasoning and fine-grained spatial grounding capture distinct dimensions of ophthalmic visual competence. Their weak co-variation supports evaluating both capabilities independently.

\subsubsection{Performance varies substantially across disease categories.}
Model performance exhibits substantial and recurring variation across disease categories, with cross-model rankings shifting according to the evaluated pathology. This trend reflects differences in disease-specific visual features, imaging characteristics, and diagnostic demands. Disease-stratified evaluation is therefore essential for characterizing category-level strengths and weaknesses beyond aggregate performance.

\section{Conclusion}

In this work, we introduce \textbf{EyeVQA}, a unified benchmark for comprehensively evaluating vision-language models in ophthalmic image understanding. 
EyeVQA integrates 21 available ophthalmic datasets and contains 20,000 question-answer pairs spanning six disease groups and seven question types, covering recognition, verification, comparative reasoning, and fine-grained spatial grounding. 
By deterministically deriving gold answers from source-provided clinical annotations and incorporating a substantial proportion of multi-image questions, EyeVQA enables reproducible and diagnostic evaluation beyond conventional single-image disease recognition.

Our systematic evaluation of representative general-purpose, scientific, and medically specialized VLMs shows that current models remain far from robust ophthalmic visual understanding. 
Although larger models generally achieve stronger overall performance, substantial variation persists across diseases and task types, while strong semantic reasoning does not necessarily translate into accurate spatial grounding. 
These findings highlight the importance of evaluating ophthalmic VLMs along multiple complementary dimensions rather than relying on a single aggregate score. 
We hope EyeVQA can serve as a standardized testbed for future research toward more reliable ophthalmic multimodal models with stronger cross-task generalization, multi-image reasoning, and fine-grained visual grounding capabilities.

\subsubsection*{Disclosure of Interests.}
The authors have no competing interests to declare that are relevant to the content of this article.

%
%
\newpage
\bibliographystyle{splncs04}
\bibliography{references}

@article{hong2025glm,
  title={Glm-4.5 v and glm-4.1 v-thinking: Towards versatile multimodal reasoning with scalable reinforcement learning},
  author={Hong, Wenyi and Yu, Wenmeng and Gu, Xiaotao and Wang, Guo and Gan, Guobing and Tang, Haomiao and Cheng, Jiale and Qi, Ji and Ji, Junhui and Pan, Lihang and others},
  journal={arXiv preprint arXiv:2507.01006},
  year={2025}
}

@article{zou2026intern,
  title={Intern-s1-pro: Scientific multimodal foundation model at trillion scale},
  author={Zou, Yicheng and Zhu, Dongsheng and Zhu, Lin and Zhu, Tong and Zhou, Yunhua and Zhou, Peiheng and Zhou, Xinyu and Zhou, Dongzhan and Zhou, Zhiwang and Zhou, Yuhao and others},
  journal={arXiv preprint arXiv:2603.25040},
  year={2026}
}

@article{xu2025lingshu,
  title={Lingshu: A generalist foundation model for unified multimodal medical understanding and reasoning},
  author={Xu, Weiwen and Chan, Hou Pong and Li, Long and Aljunied, Mahani and Yuan, Ruifeng and Wang, Jianyu and Xiao, Chenghao and Chen, Guizhen and Liu, Chaoqun and Li, Zhaodonghui and others},
  journal={arXiv preprint arXiv:2506.07044},
  year={2025}
}

@article{sellergren2025medgemma,
  title={Medgemma technical report},
  author={Sellergren, Andrew and Kazemzadeh, Sahar and Jaroensri, Tiam and Kiraly, Atilla and Traverse, Madeleine and Kohlberger, Timo and Xu, Shawn and Jamil, Fayaz and Hughes, C{\'\i}an and Lau, Charles and others},
  journal={arXiv preprint arXiv:2507.05201},
  year={2025}
}

@article{sellergren2026medgemma,
  title={Medgemma 1.5 technical report},
  author={Sellergren, Andrew and Gao, Chufan and Mahvar, Fereshteh and Kohlberger, Timo and Jamil, Fayaz and Traverse, Madeleine and Tono, Alberto and Sadjad, Bashir and Yang, Lin and Lau, Charles and others},
  journal={arXiv preprint arXiv:2604.05081},
  year={2026}
}

@article{bai2025qwen3,
  title={Qwen3-vl technical report},
  author={Bai, Shuai and Cai, Yuxuan and Chen, Ruizhe and Chen, Keqin and Chen, Xionghui and Cheng, Zesen and Deng, Lianghao and Ding, Wei and Gao, Chang and Ge, Chunjiang and others},
  journal={arXiv preprint arXiv:2511.21631},
  year={2025}
}

@inproceedings{kwon2023efficient,
  title={Efficient memory management for large language model serving with pagedattention},
  author={Kwon, Woosuk and Li, Zhuohan and Zhuang, Siyuan and Sheng, Ying and Zheng, Lianmin and Yu, Cody Hao and Gonzalez, Joseph and Zhang, Hao and Stoica, Ion},
  booktitle={Proceedings of the 29th symposium on operating systems principles},
  pages={611--626},
  year={2023}
}

@article{qin2026face,
  title={Face-human-bench: A comprehensive benchmark of face and human understanding for multi-modal assistants},
  author={Qin, Lixiong and Ou, Shilong and Zhang, Miaoxuan and Wei, Jiangning and Zhang, Yuhang and Song, Xiaoshuai and Liu, Yuchen and Wang, Mei and Xu, Weiran},
  journal={Advances in Neural Information Processing Systems},
  volume={38},
  year={2026}
}

@article{Qwen2.5-VL,
  title={Qwen2.5-VL Technical Report},
  author={Bai, Shuai and Chen, Keqin and Liu, Xuejing and Wang, Jialin and Ge, Wenbin and Song, Sibo and Dang, Kai and Wang, Peng and Wang, Shijie and Tang, Jun and Zhong, Humen and Zhu, Yuanzhi and Yang, Mingkun and Li, Zhaohai and Wan, Jianqiang and Wang, Pengfei and Ding, Wei and Fu, Zheren and Xu, Yiheng and Ye, Jiabo and Zhang, Xi and Xie, Tianbao and Cheng, Zesen and Zhang, Hang and Yang, Zhibo and Xu, Haiyang and Lin, Junyang},
  journal={arXiv preprint arXiv:2502.13923},
  year={2025}
}

@article{bazoge2026mediqal,
  title={Mediqal: A french medical question answering dataset for knowledge and reasoning evaluation},
  author={Bazoge, Adrien},
  journal={Scientific Data},
  volume={13},
  number={1},
  pages={356},
  year={2026},
  publisher={Nature Publishing Group UK London}
}

@misc{zhipuai2026glm4vflash,
  author={{GLM Team}},
  title        = {{GLM-4V-Flash}},
  howpublished = {\url{https://docs.bigmodel.cn/cn/guide/models/free/glm-4v-flash}},
  note         = {Last accessed 2026/08/11}
}

@misc{gemini3.1,
  author={{Google DeepMind}},
  title        = {{Gemini 3.1 Flash-Lite}},
  howpublished = {\url{https://ai.google.dev/gemini-api/docs/models/gemini-3.1-flash-lite}},
  note         = {Last accessed 2026/08/11}
}

@misc{gemini3.5,
  author={{Google DeepMind}},
  title        = {{Gemini 3.5 Flash-Lite}},
  howpublished = {\url{https://ai.google.dev/gemini-api/docs/models/gemini-3.5-flash-lite}},
  note         = {Last accessed 2026/08/11}
}

@misc{zhipuai2026glm41vflash,
  author={{GLM Team}},
  title        = {{GLM-4.1V-Thinking-Flash}},
  howpublished = {\url{https://docs.bigmodel.cn/cn/guide/models/vlm/glm-4.1v-thinking}},
  note         = {Last accessed 2026/08/11}
}

@misc{zhipuai2026glm46vflash,
  author={{GLM Team}},
  title        = {{GLM-4.6V-Flash}},
  howpublished = {\url{https://docs.bigmodel.cn/cn/guide/models/free/glm-4.6v-flash}},
  note         = {Last accessed 2026/08/11}
}

@misc{internlm2026interns2preview35b,
  author={{InternLM Team}},
  title        = {{Intern-S2-Preview-35B}},
  howpublished = {\url{https://huggingface.co/internlm/Intern-S2-Preview}},
  note         = {Last accessed 2026/08/11}
}

@misc{internlm2026interns2preview397b,
  author={{InternLM Team}},
  title        = {{Intern-S2-Preview-397B}},
  howpublished = {\url{https://huggingface.co/internlm/Intern-S2-Preview-397B}},
  note         = {Last accessed 2026/08/11}
}

@article{diaz2019cnns,
  title={CNNs for automatic glaucoma assessment using fundus images: an extensive validation},
  author={Diaz-Pinto, Andres and Morales, Sandra and Naranjo, Valery and K{\"o}hler, Thomas and Mossi, Jose M and Navea, Amparo},
  journal={Biomedical engineering online},
  volume={18},
  number={1},
  pages={29},
  year={2019},
  publisher={Springer}
}

@article{fang2022adam,
  title={Adam challenge: Detecting age-related macular degeneration from fundus images},
  author={Fang, Huihui and Li, Fei and Fu, Huazhu and Sun, Xu and Cao, Xingxing and Lin, Fengbin and Son, Jaemin and Kim, Sunho and Quellec, Gwenole and Matta, Sarah and others},
  journal={IEEE transactions on medical imaging},
  volume={41},
  number={10},
  pages={2828--2847},
  year={2022},
  publisher={IEEE}
}

@article{de2023airogs,
  title={Airogs: Artificial intelligence for robust glaucoma screening challenge},
  author={De Vente, Coen and Vermeer, Koenraad A and Jaccard, Nicolas and Wang, He and Sun, Hongyi and Khader, Firas and Truhn, Daniel and Aimyshev, Temirgali and Zhanibekuly, Yerkebulan and Le, Tien-Dung and others},
  journal={IEEE transactions on medical imaging},
  volume={43},
  number={1},
  pages={542--557},
  year={2023},
  publisher={IEEE}
}

@article{nakayama2024brset,
  title={BRSET: a Brazilian multilabel ophthalmological dataset of retina fundus photos},
  author={Nakayama, Luis Filipe and Restrepo, David and Matos, Jo{\~a}o and Ribeiro, Lucas Zago and Malerbi, Fernando Korn and Celi, Leo Anthony and Regatieri, Caio Saito},
  journal={PLOS Digital Health},
  volume={3},
  number={7},
  pages={e0000454},
  year={2024},
  publisher={Public Library of Science San Francisco, CA USA}
}

@article{xie2026fine,
  title={A fine-grained fundus image dataset for cataract severity assessment and diagnosis},
  author={Xie, Zixun and Ao, Mingxin and Tang, Haiming and Li, Xuemin and Bai, Xiang and Zhang, Shanghang and Li, Dawei},
  journal={Scientific Data},
  volume={13},
  number={1},
  pages={418},
  year={2026},
  publisher={Nature Publishing Group UK London}
}

@inproceedings{huang2021deepopht,
  title={Deepopht: medical report generation for retinal images via deep models and visual explanation},
  author={Huang, Jia-Hong and Yang, C-H Huck and Liu, Fangyu and Tian, Meng and Liu, Yi-Chieh and Wu, Ting-Wei and Lin, I and Wang, Kang and Morikawa, Hiromasa and Chang, Hernghua and others},
  booktitle={Proceedings of the IEEE/CVF winter conference on applications of computer vision},
  pages={2442--2452},
  year={2021}
}

@article{liu2022deepdrid,
  title={Deepdrid: Diabetic retinopathy—grading and image quality estimation challenge},
  author={Liu, Ruhan and Wang, Xiangning and Wu, Qiang and Dai, Ling and Fang, Xi and Yan, Tao and Son, Jaemin and Tang, Shiqi and Li, Jiang and Gao, Zijian and others},
  journal={Patterns},
  volume={3},
  number={6},
  year={2022},
  publisher={Elsevier}
}

@article{cuadros2009eyepacs,
  title={EyePACS: an adaptable telemedicine system for diabetic retinopathy screening},
  author={Cuadros, Jorge and Bresnick, George},
  journal={Journal of diabetes science and technology},
  volume={3},
  number={3},
  pages={509--516},
  year={2009},
  publisher={SAGE Publications}
}

@misc{kaggle_dr_rules,
  title = {Diabetic Retinopathy Detection: Competition Rules and Data Usage Terms},
  author = {{Kaggle}},
  howpublished = {\url{https://www.kaggle.com/competitions/diabetic-retinopathy-detection/rules}},
  note = {Accessed: Aug. 7, 2026}
}

@article{jin2022fives,
  title={Fives: A fundus image dataset for artificial intelligence based vessel segmentation},
  author={Jin, Kai and Huang, Xingru and Zhou, Jingxing and Li, Yunxiang and Yan, Yan and Sun, Yibao and Zhang, Qianni and Wang, Yaqi and Ye, Juan},
  journal={Scientific data},
  volume={9},
  number={1},
  pages={475},
  year={2022},
  publisher={Nature Publishing Group UK London}
}

@inproceedings{hassan2022composite,
  title={A composite retinal fundus and OCT dataset to grade macular and glaucomatous disorders},
  author={Hassan, Taimur and Raja, Hina and Hassan, Bilal and Akram, Muhammad Usman and Dias, Jorge and Werghi, Naoufel},
  booktitle={2022 2nd International Conference on Digital Futures and Transformative Technologies (ICoDT2)},
  pages={1--6},
  year={2022},
  organization={IEEE}
}

@article{wu2023gamma,
  title={Gamma challenge: glaucoma grading from multi-modality images},
  author={Wu, Junde and Fang, Huihui and Li, Fei and Fu, Huazhu and Lin, Fengbin and Li, Jiongcheng and Huang, Yue and Yu, Qinji and Song, Sifan and Xu, Xinxing and others},
  journal={Medical Image Analysis},
  volume={90},
  pages={102938},
  year={2023},
  publisher={Elsevier}
}

@article{takahashi2017applying,
  title={Applying artificial intelligence to disease staging: Deep learning for improved staging of diabetic retinopathy},
  author={Takahashi, Hidenori and Tampo, Hironobu and Arai, Yusuke and Inoue, Yuji and Kawashima, Hidetoshi},
  journal={PloS one},
  volume={12},
  number={6},
  pages={e0179790},
  year={2017},
  publisher={Public Library of Science San Francisco, CA USA}
}

@article{cen2021automatic,
  title={Automatic detection of 39 fundus diseases and conditions in retinal photographs using deep neural networks},
  author={Cen, Ling-Ping and Ji, Jie and Lin, Jian-Wei and Ju, Si-Tong and Lin, Hong-Jie and Li, Tai-Ping and Wang, Yun and Yang, Jian-Feng and Liu, Yu-Fen and Tan, Shaoying and others},
  journal={Nature communications},
  volume={12},
  number={1},
  pages={4828},
  year={2021},
  publisher={Nature Publishing Group UK London}
}

@article{qian2024competition,
  title={A competition for the diagnosis of myopic maculopathy by artificial intelligence algorithms},
  author={Qian, Bo and Sheng, Bin and Chen, Hao and Wang, Xiangning and Li, Tingyao and Jin, Yixiao and Guan, Zhouyu and Jiang, Zehua and Wu, Yilan and Wang, Jinyuan and others},
  journal={JAMA ophthalmology},
  volume={142},
  number={11},
  pages={1006--1015},
  year={2024}
}

@inproceedings{li2020benchmark,
  title={A benchmark of ocular disease intelligent recognition: One shot for multi-disease detection},
  author={Li, Ning and Li, Tao and Hu, Chunyu and Wang, Kai and Kang, Hong},
  booktitle={International symposium on benchmarking, measuring and optimization},
  pages={177--193},
  year={2020},
  organization={Springer}
}

@inproceedings{zhang2010origa,
  title={Origa-light: An online retinal fundus image database for glaucoma analysis and research},
  author={Zhang, Zhuo and Yin, Feng Shou and Liu, Jiang and Wong, Wing Kee and Tan, Ngan Meng and Lee, Beng Hai and Cheng, Jun and Wong, Tien Yin},
  booktitle={2010 Annual international conference of the IEEE engineering in medicine and biology},
  pages={3065--3068},
  year={2010},
  organization={IEEE}
}

@article{fang2024open,
  title={Open fundus photograph dataset with pathologic myopia recognition and anatomical structure annotation},
  author={Fang, Huihui and Li, Fei and Wu, Junde and Fu, Huazhu and Sun, Xu and Orlando, Jos{\'e} Ignacio and Bogunovi{\'c}, Hrvoje and Zhang, Xiulan and Xu, Yanwu},
  journal={Scientific data},
  volume={11},
  number={1},
  pages={99},
  year={2024},
  publisher={Nature Publishing Group UK London}
}

@article{benitez2021dataset,
  title={Dataset from fundus images for the study of diabetic retinopathy},
  author={Ben{\'\i}tez, Veronica Elisa Castillo and Matto, Ingrid Castro and Rom{\'a}n, Julio C{\'e}sar Mello and Noguera, Jos{\'e} Luis V{\'a}zquez and Garc{\'\i}a-Torres, Miguel and Ayala, Jordan and Pinto-Roa, Diego P and Gardel-Sotomayor, Pedro E and Facon, Jacques and Grillo, Sebastian Alberto},
  journal={Data in brief},
  volume={36},
  pages={107068},
  year={2021},
  publisher={Elsevier}
}

@article{fang2022refuge2,
  title={Refuge2 challenge: A treasure trove for multi-dimension analysis and evaluation in glaucoma screening},
  author={Fang, Huihui and Li, Fei and Wu, Junde and Fu, Huazhu and Sun, Xu and Son, Jaemin and Yu, Shuang and Zhang, Menglu and Yuan, Chenglang and Bian, Cheng and others},
  journal={arXiv preprint arXiv:2202.08994},
  year={2022}
}

@article{niemeijer2009retinopathy,
  title={Retinopathy online challenge: automatic detection of microaneurysms in digital color fundus photographs},
  author={Niemeijer, Meindert and Van Ginneken, Bram and Cree, Michael J and Mizutani, Atsushi and Quellec, Gw{\'e}nol{\'e} and S{\'a}nchez, Clara I and Zhang, Bob and Hornero, Roberto and Lamard, Mathieu and Muramatsu, Chisako and others},
  journal={IEEE transactions on medical imaging},
  volume={29},
  number={1},
  pages={185--195},
  year={2009},
  publisher={IEEE}
}

@article{liu2025human,
  title={Human-MME: A holistic evaluation benchmark for human-centric multimodal large language models},
  author={Liu, Yuansen and Tang, Haiming and Peng, Jinlong and Zhang, Jiangning and Ji, Xiaozhong and He, Qingdong and Wu, Wenbin and Luo, Donghao and Gan, Zhenye and Zhu, Junwei and others},
  journal={arXiv preprint arXiv:2509.26165},
  year={2025}
}

@article{wang2026towards,
  title={Towards Clinically Interpretable Ophthalmic VQA via Spatially-Grounded Lesion Evidence},
  author={Wang, Xingyue and Liu, Bo and Wang, Meng and Zhang, Zhixuan and Zhu, Chengcheng and Fu, Huazhu and Liu, Jiang},
  journal={arXiv preprint arXiv:2605.22414},
  year={2026}
}

@inproceedings{goyal2017making,
  title={Making the v in vqa matter: Elevating the role of image understanding in visual question answering},
  author={Goyal, Yash and Khot, Tejas and Summers-Stay, Douglas and Batra, Dhruv and Parikh, Devi},
  booktitle={Proceedings of the IEEE conference on computer vision and pattern recognition},
  pages={6904--6913},
  year={2017}
}

@inproceedings{hudson2019gqa,
  title={Gqa: A new dataset for real-world visual reasoning and compositional question answering},
  author={Hudson, Drew A and Manning, Christopher D},
  booktitle={2019 IEEE/CVF Conference on Computer Vision and Pattern Recognition (CVPR)},
  pages={6693--6702},
  year={2019},
  organization={IEEE}
}

@article{lau2018dataset,
  title={A dataset of clinically generated visual questions and answers about radiology images},
  author={Lau, Jason J and Gayen, Soumya and Ben Abacha, Asma and Demner-Fushman, Dina},
  journal={Scientific data},
  volume={5},
  number={1},
  pages={180251},
  year={2018},
  publisher={Nature Publishing Group}
}

@article{he2020pathvqa,
  title={Pathvqa: 30000+ questions for medical visual question answering},
  author={He, Xuehai and Zhang, Yichen and Mou, Luntian and Xing, Eric and Xie, Pengtao},
  journal={arXiv preprint arXiv:2003.10286},
  year={2020}
}

@article{zhou2023foundation,
  title={A foundation model for generalizable disease detection from retinal images},
  author={Zhou, Yukun and Chia, Mark A and Wagner, Siegfried K and Ayhan, Murat S and Williamson, Dominic J and Struyven, Robbert R and Liu, Timing and Xu, Moucheng and Lozano, Mateo G and Woodward-Court, Peter and others},
  journal={Nature},
  volume={622},
  number={7981},
  pages={156--163},
  year={2023},
  publisher={Nature Publishing Group UK London}
}

@article{silva2025foundation,
  title={A foundation language-image model of the retina (flair): Encoding expert knowledge in text supervision},
  author={Silva-Rodriguez, Julio and Chakor, Hadi and Kobbi, Riadh and Dolz, Jose and Ayed, Ismail Ben},
  journal={Medical Image Analysis},
  volume={99},
  pages={103357},
  year={2025},
  publisher={Elsevier}
}

@misc{eyepacs,
  author = {{Kaggle}},
  title = {{Diabetic Retinopathy Detection Challenge}},
  howpublished = "\url{https://www.kaggle.com/c/diabetic-retinopathy-detection}",
  note = {Last accessed 2026/8/11}
}

@inproceedings{li2019attention,
  title={Attention based glaucoma detection: A large-scale database and CNN model},
  author={Li, Liu and Xu, Mai and Wang, Xiaofei and Jiang, Lai and Liu, Hanruo},
  booktitle={2019 IEEE/CVF Conference on Computer Vision and Pattern Recognition (CVPR)},
  pages={10563--10572},
  year={2019},
  organization={IEEE}
}

@article{pachade2021retinal,
  title={Retinal fundus multi-disease image dataset (rfmid): A dataset for multi-disease detection research},
  author={Pachade, Samiksha and Porwal, Prasanna and Thulkar, Dhanshree and Kokare, Manesh and Deshmukh, Girish and Sahasrabuddhe, Vivek and Giancardo, Luca and Quellec, Gwenol{\'e} and M{\'e}riaudeau, Fabrice},
  journal={Data},
  volume={6},
  number={2},
  pages={14},
  year={2021},
  publisher={MDPI}
}

@article{orlando2020refuge,
  title={Refuge challenge: A unified framework for evaluating automated methods for glaucoma assessment from fundus photographs},
  author={Orlando, Jos{\'e} Ignacio and Fu, Huazhu and Breda, Jo{\~a}o Barbosa and Van Keer, Karel and Bathula, Deepti R and Diaz-Pinto, Andr{\'e}s and Fang, Ruogu and Heng, Pheng-Ann and Kim, Jeyoung and Lee, JoonHo and others},
  journal={Medical image analysis},
  volume={59},
  pages={101570},
  year={2020},
  publisher={Elsevier}
}

@misc{choi2021retina_dataset,
 author = {Sungjoon Choi},
 title = {retina\_dataset: Fundus Image Dataset for Eye Disease Classification},
 year = {2021},
 publisher = {GitHub},
 journal = {GitHub repository},
 howpublished = {\url{https://github.com/yiweichen04/retina_dataset}},
 note = {Accessed: Aug. 7, 2026}
}

%
%
\newpage
\clearpage
\appendix

\begin{center}
  \vspace*{1em}
  {\LARGE\bfseries Supplementary Material for EyeVQA\par}
  \vspace{1.5em}
\end{center}

\titlerunning{Supplementary Material: EyeVQA}

\renewcommand{\thefigure}{S\arabic{figure}}
\renewcommand{\thetable}{S\arabic{table}}
\renewcommand{\thesection}{\Alph{section}}
\renewcommand{\thesubsection}{\Alph{section}.\arabic{subsection}}
\setcounter{section}{0}
\setcounter{table}{0}
\setcounter{figure}{0}

\section{Overview}
\label{supp:intro}
This supplementary material complements the main text of EyeVQA by providing additional technical details and formal specifications. Specifically, Section~\ref{supp:dataset_collection} details the collection, institutional origins, disease coverage, and preprocessing procedures of the 21 ophthalmic datasets summarized in Table~\ref{tab:datasets}. Section~\ref{supp:qa_details} elaborates on the design protocols, distractor sampling strategies, and coordinate system conventions across all seven standardized question types. Section~\ref{supp:evaluation_metrics} presents the explicit mathematical formulations of the evaluation metrics, score normalization rules, and refusal handling mechanisms.

\section{Supplementary Details for Dataset Collection}
\label{supp:dataset_collection}

\subsection{Public Ophthalmic Dataset Collection}
The rapid development of ophthalmic image analysis has led to the release of numerous retinal image datasets over the past decade. Most of these datasets were designed for individual diseases or specific clinical applications, such as diabetic retinopathy grading, glaucoma screening, pathological myopia diagnosis, and retinal lesion segmentation. Although these datasets have significantly promoted the development of computer-aided diagnosis systems, few single datasets provide sufficient disease diversity or annotation richness for comprehensively evaluating ophthalmic vision-language models (VLMs). Therefore, instead of relying on a single data source, we collected multiple available ophthalmic datasets to support the construction of a comprehensive ophthalmic VLM benchmark.

The final collection consists of \textbf{21 available ophthalmic datasets} covering major retinal diseases, including diabetic retinopathy, glaucoma, pathological myopia, macula disease, cataract, retinal vascular abnormalities, and other common retinal disorders. We collected these datasets from diverse sources, including \textit{Scientific Data} journal publications, Kaggle challenges, Hugging Face datasets, Figshare, Zenodo, PhysioNet, Mendeley Data, CodaLab platforms, Grand Challenge platforms, and GitHub repositories. These datasets have been widely adopted in ophthalmic artificial intelligence research. An overview of all collected datasets is presented in Table~\ref{tab:datasets}.

The collected datasets exhibit substantial diversity in both clinical content and annotation protocols. Disease-specific datasets mainly focus on a single ophthalmic disorder with carefully designed grading criteria or lesion annotations, whereas multi-disease datasets include multiple retinal abnormalities within a unified labeling framework. Besides image-level disease classification, several datasets provide anatomical structure segmentation, retinal vessel segmentation, lesion localization, paired fundus--OCT images, demographic information, and clinical descriptions. Such heterogeneous annotations enable the construction of evaluation tasks ranging from visual perception to clinically oriented reasoning, providing a much more comprehensive assessment of ophthalmic vision-language models than conventional image classification benchmarks.

Another important characteristic of the collected datasets is their diversity in imaging conditions. Since the datasets were collected by different medical institutions using various fundus cameras under different clinical environments, considerable differences exist in image resolution, field of view, illumination, image quality, annotation standards, and metadata organization. These discrepancies increase the difficulty of cross-dataset evaluation while better reflecting real-world clinical scenarios, thereby improving the robustness and generalizability of the benchmark.

To ensure consistency during benchmark construction, all datasets were standardized before downstream processing. Image formats, annotation schemas, disease terminology, metadata fields, and directory structures were unified through a preprocessing pipeline, thereby reducing inconsistencies introduced by heterogeneous data sources. The standardized datasets were subsequently used for automatic question--answer generation.

\begin{table}[htbp]
\caption{Overview of the collected ophthalmic datasets.}
\label{tab:datasets}
\scriptsize
\centering
\begin{tabular}{|p{2.0cm}|r|p{4cm}|p{2.0cm}|p{2.0cm}|}
\hline
\textbf{Dataset} & \textbf{Images} & \textbf{Source (institution/country)} & \textbf{Disease} & \textbf{License} \\
\hline
ACRIMA \cite{diaz2019cnns} \newline (2019) & 705 & FISABIO Oftalmolog\'ia M\'edica, Valencia, Spain & Glaucoma & CC BY 4.0 \\
\hline
ADAM \cite{fang2022adam} \newline (2018) & 1,200 & Zhongshan Ophthalmic Center, Sun Yat-sen University, China & Macula Disease & CC BY-NC-ND 4.0 \\
\hline
AIROGS \cite{de2023airogs} \newline (2021) & 101,442 & 500 screening centers across the United States of America & Glaucoma & CC BY-NC-ND 4.0 \\
\hline
BRSET \cite{nakayama2024brset} \newline (2024) & 16,266 & Department of Ophthalmology, S\~ao Paulo Federal University, Brazil & Diabetic Retinopathy, Macula Disease, etc. & PhysioNet Credentialed Health Data License 1.5.0 \\
\hline
CSDI \cite{xie2026fine} \newline (2026) & 187 & Peking University Third Hospital, China & Cataract & CC BY 4.0 \\
\hline
Retina Dataset \cite{choi2021retina_dataset} \newline (2016) & 601 & 8 open access databases (specific sources undisclosed) & Glaucoma, Cataract, etc. & Undisclosed \\
\hline
DEN \cite{huang2021deepopht} \newline (2021) & 15,708 & Undisclosed (NDA required) & Diabetic Retinopathy, Glaucoma, Macula Disease, etc. & Undisclosed \\
\hline
DeepDRiD \cite{liu2022deepdrid} \newline (2022) & 2,256 & Department of Ophthalmology, Shanghai Jiao Tong University Affiliated Sixth People's Hospital, Shanghai, China & Diabetic Retinopathy & CC BY-SA 4.0 \\
\hline
EyePACS Dataset \cite{cuadros2009eyepacs,kaggle_dr_rules} \newline (2015) & 88,702 & EyePACS platform (USA) and 3 eye hospitals in India (Aravind Eye Hospital, Sankara Nethralaya, Narayana Nethralaya) & Diabetic Retinopathy & Kaggle Competition Rules \\
\hline
FIVES \cite{jin2022fives} \newline (2022) & 800 & Ophthalmology Centre at the Second Affiliated Hospital of Zhejiang University, China & Diabetic Retinopathy, Glaucoma, Macula Disease, etc. & CC BY 4.0 \\
\hline
FUND-OCT \cite{hassan2022composite} \newline (2021) & 200 & Undisclosed & Glaucoma, Macula Disease, etc. & CC BY 4.0 \\
\hline
GAMMA \cite{wu2023gamma} \newline (2021) & 300 & Zhongshan Ophthalmic Center, Sun Yat-sen University, Guangzhou, China & Glaucoma & CC BY-NC-ND \\
\hline
JICHI Dataset \cite{takahashi2017applying} \newline (2017) & 9,940 & Jichi Medical University Hospital, Tochigi, Japan & Diabetic Retinopathy & CC BY 4.0 \\
\hline
JSIEC1000 \cite{cen2021automatic} \newline (2019) & 1,000 & Joint Shantou International Eye Centre (JSIEC), Shantou, Guangdong, China & Diabetic Retinopathy, Glaucoma, Macula Disease, etc. & DbCL v1.0 \\
\hline
MMAC \cite{qian2024competition} \newline (2023) & 1,391 & Shanghai Health and Medical Center (SHMC), China; Shanghai Sixth People's Hospital (SSPH), China & Pathological Myopia & CC BY 4.0 \\
\hline
ODIR \cite{li2020benchmark} \newline (2019) & 8,000 & Multiple hospitals and medical centers in China & Diabetic Retinopathy, Glaucoma, Cataract, Macula Disease, etc. & Undisclosed \\
\hline
ORIGA-light \cite{zhang2010origa} \newline (2010) & 650 & Singapore Eye Research Institute, Singapore & Glaucoma & Undisclosed \\
\hline
PALM \cite{fang2024open} \newline (2019) & 1,200 & Zhongshan Ophthalmic Center, Sun Yat-sen University, Guangzhou, China & Pathological Myopia & Non-redistributable \\
\hline
PARAGUAY \cite{benitez2021dataset} \newline (2021) & 1,437 & Department of Ophthalmology, Hospital de Cl\'inicas, Facultad de Ciencias M\'edicas, Universidad Nacional de Asunci\'on, San Lorenzo, Paraguay & Diabetic Retinopathy & CC BY 4.0 \\
\hline
REFUGE2 \cite{orlando2020refuge,fang2022refuge2} \newline (2020) & 1,200 & Multiple medical centers (specific sources not fully disclosed) & Glaucoma & Non-redistributable \\
\hline
ROC \cite{niemeijer2009retinopathy} \newline (2009) & 100 & Undisclosed (from a large diabetic retinopathy screening program) & Diabetic Retinopathy & Non-redistributable \\
\hline
\end{tabular}
\end{table}

\subsection{Data Distribution and Licensing Policy}
\label{supp:licensing_policy}

To respect the intellectual property rights and raw data terms of service of the original dataset creators, EyeVQA strictly adheres to a dual-track data distribution policy based on the underlying dataset licenses (detailed in Table~\ref{tab:datasets}):

\begin{enumerate}
    \item \textbf{Direct Image Redistribution (Permissive Licenses):} For datasets released under permissive open-access licenses that explicitly allow redistribution and adaptation (e.g., CC BY 4.0, CC BY-SA 4.0, such as CSDI~\cite{xie2026fine}, ACRIMA~\cite{diaz2019cnns}, and FIVES~\cite{jin2022fives}), we directly distribute the standardized fundus images alongside our generated QA pairs in the public benchmark repository for maximum ease of evaluation.
    
    \item \textbf{Metadata-Only Distribution (Restrictive / Non-Redistributable Licenses):} For datasets governed by restrictive licenses, NDA requirements, or non-redistributable terms (e.g., CC BY-NC-ND 4.0, PhysioNet Credentialed Licenses, Kaggle Competition Rules, or proprietary terms, such as ADAM~\cite{fang2022adam}, BRSET~\cite{nakayama2024brset}, EyePACS~\cite{cuadros2009eyepacs}, and PALM~\cite{fang2024open}), \textbf{we do not host or redistribute the raw fundus images}. Instead, we release only the deterministic QA pairs mapped to the original image filenames/identifiers. Users can obtain the raw images directly from the official repositories via the provided references/URLs and align them with our VQA benchmark using our provided automated pairing scripts.
\end{enumerate}

This protocol ensures full compliance with medical data governance and original licensing restrictions while providing a standardized, reproducible evaluation framework for the community.

\section{Detailed Question--Answer Generation Guidelines}
\label{supp:qa_details}

This section provides comprehensive specifications for the seven standardized question types in EyeVQA, including distractor sampling, ordering criteria, coordinate system conventions, and metric definitions.

\paragraph{Single-Choice (SC), Multi-Select (MS), and Variable-Select (VS) Questions.}
SC questions present exactly four candidate options (A--D) with one deterministic gold answer, testing disease identification, severity grading, or image--text correspondence. MS questions explicitly specify the requirement of multiple correct options. VS questions leave the answer cardinality unstated, requiring models to dynamically determine whether one or several options are correct. 
To construct challenging distractors and mitigate language priors~\cite{goyal2017making,hudson2019gqa}, incorrect options are preferentially drawn from clinically related disease classes, adjacent severity levels, or visually similar cases within the same dataset.

\paragraph{True-False (TF) and Ranking (RK) Questions.}
TF questions prompt the model to verify whether a given clinical statement, diagnostic claim, or qualitative descriptor matches the visual evidence. 
RK questions evaluate ordinal perception by requiring the model to sort a sequence of fundus images according to a designated clinical attribute (e.g., disease severity, lesion burden, or cup-to-disc ratio). To avoid ambiguity, RK instances are included only when ground-truth ordering is strictly monotonic and clinically indisputable.

\paragraph{Point Location (PL) and Bounding Box (BB) Questions.}
PL and BB assess spatial perception and visual grounding in fundus photography:
\begin{itemize}
    \item \textbf{Point Location (PL):} Predicts a single normalized coordinate
    \[
    [x, y] \in [0, 1]^2
    \]
    representing key anatomical landmarks, such as the optic disc center or macula fovea.
    \item \textbf{Bounding Box (BB):} Predicts a normalized bounding box
    \[
    [x_{\min}, y_{\min}, x_{\max}, y_{\max}] \in [0, 1]^4
    \]
    circumscribing specific lesions or structural entities.
\end{itemize}
All spatial targets are extracted from expert-annotated segmentation masks, boxes, or landmarks. During dataset preprocessing, target coordinates undergo exact geometric transformations aligned with image cropping and resizing; samples with inconsistent or ambiguous spatial annotations are systematically excluded~\cite{wang2026towards}.

\section{Detailed Formulations of Evaluation Metrics}
\label{supp:evaluation_metrics}

Let $N_t$ denote the number of questions of type $t$. Metrics are reported separately for each question type, with invalid responses assigned a value of zero.

\subsection{Single-Choice and True-False Questions}
We evaluate $\mathrm{SC}$ and $\mathrm{TF}$ questions using exact-match accuracy:
\begin{equation} 
z_i=\left\{\begin{array}{ll} 1, & \hat{y}_i=y_i,\\ 0, & \mathrm{otherwise}, \end{array}\right. \qquad \mathrm{Acc}_{t}=\frac{1}{N_t}\sum_{i=1}^{N_t}z_i, \quad t\in\{\mathrm{SC},\mathrm{TF}\}. 
\label{eq:discrete_accuracy} 
\end{equation}
where $y_i$ and $\hat{y}_i$ are the reference and parsed answers, respectively.

\subsection{Multi-Select and Variable-Select Questions}
We evaluate $\mathrm{MS}$ and $\mathrm{VS}$ questions using the Hamming metric (HM) \cite{bazoge2026mediqal}:
\begin{equation} 
\mathrm{HM}_{t}=\frac{1}{N_t}\sum_{i=1}^{N_t}\frac{|Y_i\cap\hat{Y}_i|}{|Y_i\cup\hat{Y}_i|}. 
\label{eq:hm_score} 
\end{equation}
where $t\in\{\mathrm{MS},\mathrm{VS}\}$, and $Y_i$ and $\hat{Y}_i$ are the reference and predicted option sets, respectively. The metric assigns partial credit according to set overlap.

\subsection{Ranking Questions}
We evaluate $\mathrm{RK}$ questions using Kendall's Tau \cite{qin2026face}:
\begin{equation} 
\mathrm{Tau}=\frac{1}{N_t}\sum_{i=1}^{N_t}\frac{C_i-D_i}{K(K-1)/2}. 
\label{eq:kendall_tau} 
\end{equation}
where $K=4$ is the number of ranked options, and $C_i$ and $D_i$ are the number of concordant and discordant pairs among $K$ items.

\subsection{Bounding Box Questions}
We evaluate $\mathrm{BB}$ questions using intersection over union (IoU):
\begin{equation} 
\mathrm{IoU}=\frac{1}{N_t}\sum_{i=1}^{N_t}\frac{\mathrm{Area}(B_i\cap\hat{B}_i)}{\mathrm{Area}(B_i\cup\hat{B}_i)}. 
\label{eq:iou} 
\end{equation}
where $B_i$ is normalized reference box and $\hat{B}_i$ is normalized predicted box.

\subsection{Point Location Questions}
We evaluate $\mathrm{PL}$ questions using accuracy with a normalized Euclidean-distance tolerance of $0.05$:
\begin{equation} 
\begin{array}{c} 
z_i^{\mathrm{L}}=\left\{\begin{array}{ll} 1, & \sqrt{(\hat{x}_i-x_i)^2+(\hat{y}_i-y_i)^2}\leq0.05,\\ 0, & \mathrm{otherwise}, \end{array}\right.\\[6pt] 
\mathrm{Acc}_{\mathrm{L}@0.05}=\displaystyle\frac{1}{N_t}\sum_{i=1}^{N_t}z_i^{\mathrm{L}}. 
\end{array} 
\label{eq:localization_accuracy} 
\end{equation}
where $p_i=(x_i,y_i)$ and $\hat{p}_i=(\hat{x}_i,\hat{y}_i)$ are the normalized reference and predicted points, respectively.

\paragraph{Score Normalization and Refusal Handling.}
For comparison, all metrics are linearly normalized so that the final scores lie in $[0,100]$: values in $[0,1]$ are multiplied by $100$, while Kendall's Tau is mapped from $[-1,1]$ to $[0,100]$. If a VLM refuses to answer a question, that question receives a score of zero and remains included in the average.

\end{document}